\documentclass{bmvc2k}

\usepackage{booktabs}
\usepackage{multirow}
\usepackage[toc,page]{appendix}

\title{Tripping through time: Efficient Localization of Activities in Videos}

\addauthor{Meera Hahn}{meerahahn@gatech.edu}{1}
\addauthor{Asim Kadav}{asim@nec-labs.com}{2}
\addauthor{James M. Rehg}{rehg@gatech.edu}{1}
\addauthor{Hans Peter Graf}{hpg@nec-labs.com}{2}

\addinstitution{
 College of Computing\\
 Georgia Institute of Technology\\
 Atlanta, GA
}
\addinstitution{
 Machine Learning Department\\
 NEC Labs America\\
 Princeton, NJ
}

\runninghead{Hahn, Kadav, Rehg, Graf}{Efficient Video Localization}

\begin{document}
\maketitle
\begin{abstract}
Localizing moments in untrimmed videos via language queries is a new and interesting task that requires the ability to accurately ground language into video. Previous works have approached this task by processing the entire video, often more than once, to localize relevant activities. In the real world applications of this approach, such as video surveillance, efficiency is a key system requirement. In this paper, we present TripNet, an end-to-end system that uses a gated attention architecture to model fine-grained textual and visual representations in order to align text and video content. Furthermore, TripNet uses reinforcement learning to efficiently localize relevant activity clips in long videos, by learning how to intelligently skip around the video. It extracts visual features for few frames to perform activity classification. In our evaluation over Charades-STA~\cite{gao2017tall}, ActivityNet Captions~\cite{krishna2017dense} and the TACoS dataset~\cite{rohrbach2012database}, we find that TripNet achieves high accuracy and saves processing time by only looking at 32-41\% of the entire video. 
\end{abstract}
\section{Introduction}
The increasing availability of videos and their importance in application domains, such as social media and surveillance, has created a pressing need for automated video analysis methods. A particular challenge arises in long video clips which have noisy or nonexistent labels, which are common in many settings including surveillance and instructional videos. While classical video retrieval works at the level of entire clips, a more challenging and important task is to efficiently sift through large amounts of unorganized video content and retrieve specific moments of interest. 

\begin{figure}[t]
    \centering
    \includegraphics[width=3.1in]{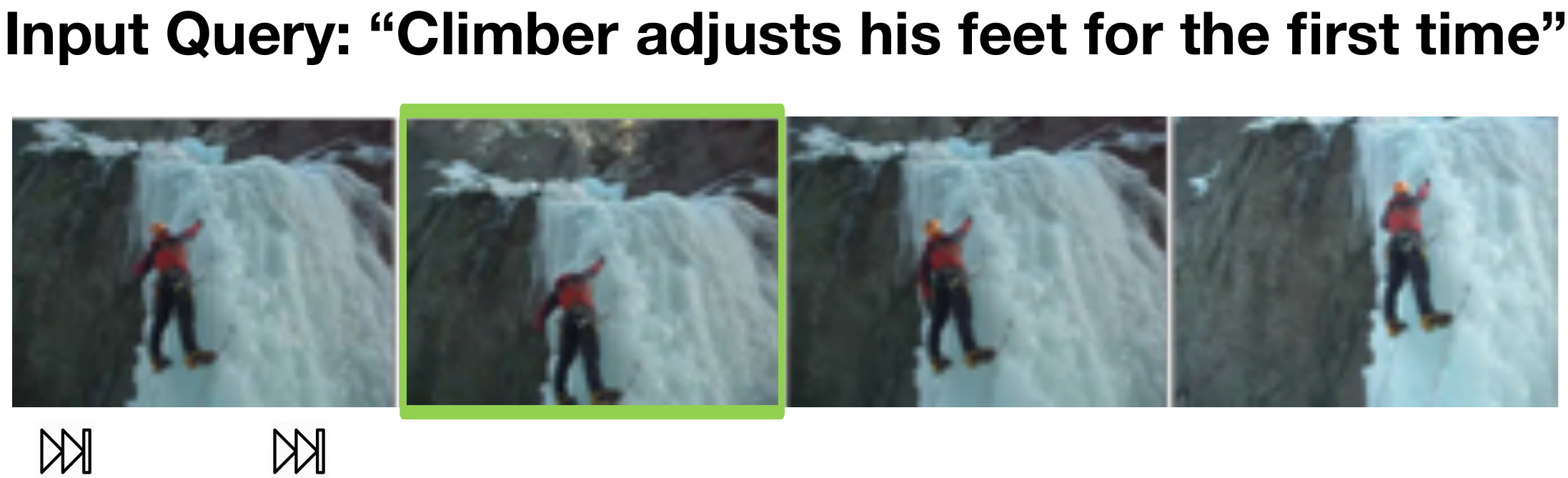}
    \caption {Figure demonstrating video search method where given an input query, a human would normally forward the video until relevant objects or objects arrive and then proceed slowly to obtain the appropriate temporal boundaries.}
    \label{fig:howtotrip}
    \vspace{-0.2in}
\end{figure}

A promising formulation of this task is the paradigm of temporal activity localization via language query (TALL), which was introduced by Gao et al.~\cite{gao2017tall} and Hendricks et al.~\cite{hendricks2017localizing}. The TALL task is illustrated in Fig.~\ref{fig:howtotrip} and shows a few frames from a climbing video along with a language query describing the event of interest ``Climber adjusts his feet for the first time.'' In the figure the green frame denotes the desired output, the first frame in which the feet are being adjusted. Note that the solution to this problem requires local and global temporal information: local cues are needed to identify the specific frame in which the feet are adjusted, but global cues are also needed to identify the first time this occurs, since the climber adjusts his feet throughout the clip. 
The MCN~\cite{hendricks2017localizing} and CTRL~\cite{gao2017tall} architectures, along with more recent work by Yuan et al.~\cite{yuan2018find}, have obtained encouraging results for the TALL task, however, they all suffer from a significant limitation which we address in this paper: Existing methods construct temporally-dense representations of the entire video clip which are then analyzed to identify the target events. This can be very inefficient in long video clips, where the target might be a single short moment. This approach starkly contrasts to how humans search for events of interest, as illustrated schematically in Fig.~\ref{fig:howtotrip}. A human would fast-forward through the video from the beginning, effectively sampling sparse sets of frames until they got closer to the region of interest. Then, they would look frame-by-frame until the start and end points are localized. At that point the search would terminate, leaving the vast majority of frames unexamined. Note that efficient solutions must go well beyond simple heuristics, since events of interest can occur anywhere within a target clip and the global position may not be obvious from the query.

In order to develop an efficient solution to the TALL task it is necessary to address two main challenges: 1) Devising an effective joint representation (embedding) of video and language features to support localization; and 2) Learning an efficient search strategy that can mimic the human ability to sample a video clip intelligently to localize an event of interest. Most prior works on TALL address the issue of constructing a joint embedding space through a sliding window approach that pools at different temporal scales. This strategy has achieved some success, but it is not well suited for efficient search because it provides only a coarse control over which frames are evaluated. In contrast, our usage of gated-attention architecture is effective in aligning the text queries, which  often consist of an object and its attributes or an action, with the video features that consist of convolutional filters that can identify these elements. Two prior works~\cite{yuan2018find,xu2018text} also utilize an attention model, but they do not address its use in efficient search. 

We address the challenge of efficient search through a combination of reinforcement learning (RL) and fine-grained video analysis. Our approach is inspired in part by strategies used by human video annotators. In addition, we make an analogy to the success of RL in language-guided navigation tasks in 3D environments~\cite{anderson2018vision, chaplot2018gated, das2018embodied, gordon2018iqa}. Specifically, we make an analogy between temporally localizing events in a video through playback controls and the actions an agent would take to navigate around an environment looking for a specific object of interest. We share with the navigation task, the fact that labeled data is not available for explicitly modeling the relationship between actions and rewards, but it is possible to learn a model through simulation. Our approach to temporal localization uses a novel architecture for combining the multi-modal video and text features with a policy learning module that learns to step forward and rewind the video and receives awards for accurate temporal localization.

In summary this paper makes two contributions: First, we present a novel end-to-end reinforcement learning framework called TripNet that addresses the problem of temporal activity localization via a language query. TripNet uses gated-attention to align text and visual features, leading to improved accuracy. Second, we present experimental results on the datasets Charades-STA~\cite{gao2017tall}, ActivityNet Captions~\cite{krishna2017dense} and  TACoS~\cite{rohrbach2012database}. These results demonstrate that TripNet achieves state of the art results in accuracy while significantly increasing efficiency, by evaluating only 32-41\% of the total video.
\section{Related Work}
\paragraph{Querying using Natural Language}
This paper is most-closely related to prior works on the TALL problem, beginning with the two works~\cite{gao2017tall,hendricks2017localizing} that introduced it. These works additionally introduced a dataset for the task, Charades-STA\cite{gao2017tall} and DiDeMo\cite{hendricks2017localizing}. Each dataset contains untrimmed videos with multiple sentence queries and the corresponding start and end timestamp of the clip within the video. The two papers adopt a supervised cross modal embedding approach, in which sentences and videos are projected into a embedding space, optimized so that queries and their corresponding video clips will lie close together, while non-corresponding clips will be far apart. At test time both approaches run a sliding window across the video and compute an alignment score between the candidate window and the language query. The window with the highest score is then selected as the localized query. Follow-up works on the TALL task have differed in the design of the embedding process~\cite{chen2018temporally}. \cite{liu2018attentive,song2018val,yuan2018find} modified the original approach by adding self-attention and co-attention to the embedding process. \cite{xu2018text} introduces early fusion of the text queries and video features rather than using an attention mechanism. Additionally, \cite{xu2018text} uses the text to produce activity segment proposals as their candidate windows instead of using a fixed sliding window approach. Pre-processing of the sentence queries has not been a primary focus of previous works, with most methods using a simple LSTM based architecture for sentence embedding, with  Glove~\cite{pennington2014glove} or Skip-Thought~\cite{kiros2015skip} vectors to represent the sentences. Other work in video-based localization which predated TALL either uses a limited set of text vocabulary to search for specific events or uses structured videos~\cite{bojanowski2015weakly,sener2015unsupervised,tellex2009towards,hahn2018learning}. Other earlier works address the task of retrieving objects using natural language questions, which is less complex than than localizing actions in videos~\cite{hu2016natural,johnson2017clevr,mao2016generation}.
\vspace{-0.1in}
\paragraph{Temporal Localization}
Temporal action localization refers to localizing activities over a known set of action labels in untrimmed videos. Some existing work in this area has found success by extracting CNN features from the video frames, pooling the features and feeding them into either single or multi-stage classifiers to obtain action predictions, along with temporal labels~\cite{chao2018rethinking,dai2017temporal,yuan2016temporal,xu2017r}. These methods use a sliding window approach. Often, multiple window sizes are used and candidate windows are densely sampled, meaning that the candidates overlap with each other. This method has high accuracy, but it is an exhaustive search method that leads to high computational costs. There are also TALL methods that forgo the end-to-end approach and instead use a two-stage approach: first generating temporal proposals, and second, classifying the proposals into action categories~\cite{buch2017sst, gao2017cascaded, ghosh2019excl, ge2019mac, heilbron2017scc, shou2017cdc, rodriguez2020proposal, shou2016temporal}. Unlike both sets of previous methods, our proposed model TripNet, uses reinforcement learning to perform temporal localization using natural language queries. In action-recognition methods~\cite{bian2017revisiting, ma2018attend, donahue2015long, liu2009recognizing, miech2017learnable,sigurdsson2016asynchronous, tran2014c3d}, Frame glimpse~\cite{yeung2016end} and Action Search~\cite{alwassel2018action} use reinforcement learning to identify an action while looking at the smallest possible number of frames. These works focus only on classifying actions (as opposed to parsing natural language activities) in trimmed videos and and the former does not perform any type of temporal boundary localization. A recent work~\cite{he2019read} also uses RL to learn the temporal video boundaries. However, it analyzes the entire video and does not focus on the efficiency of localizing the action. Another work~\cite{wang2019language}, uses RL to estimate the activity location using fixed temporal boundaries. Furthermore, it uses Faster R-CNN trained on the Visual Genome dataset to provide semantic concepts to the model. 
\vspace{-0.1in}
\paragraph{3D navigation and Reinforcement Learning}
Locating a specific activity within a video using a reinforcement learning agent is analogous to the task of navigating through a three dimensional world. How can we efficiently learn to navigate through the temporal landscape of the video? The parallels between our approach and navigation lead us to additionally review the related work in the areas of language-based navigation and embodied perception~\cite{anderson2018vision,gordon2018iqa}. Recently, the task of Embodied Question Answering~\cite{das2018embodied} was introduced. In this task, an agent is asked questions about the environment, such as ``what color is the bathtub,'' and the agent must navigate to the bathtub and answer the question. The method introduced in \cite{das2018embodied} focuses on grounding the language of the question not directly into the pixels of the scene but into actions for navigating around the scene. We seek to ground our text query into actions for skipping around the video to narrow down on the correct clip. Another recent work ~\cite{chaplot2018gated} explores giving visually grounded instructions such as ``go to the green red torch'' to an agent and having them learn to navigate their environment to find the object and complete the instruction. 
\section{Methods}
\begin{figure*}[t]
    \centering
    \includegraphics[width=\textwidth]{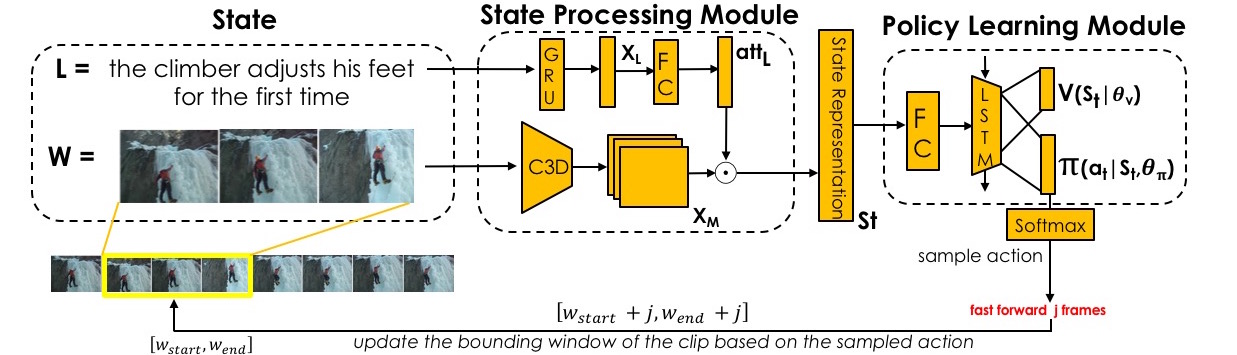}
    \caption{TripNet localizes specific moments in videos based on a natural language query. There are two main components of TripNet: the state processing module and the policy learning module. The state processing module encodes the state into a joint visual and linguistic representation which is then fed to the policy learning module on which it generates the action policy. The sampled action in this figure is skipping forward by j frames and the state is updated accordingly.}
    \label{fig:tripnet-model}
    \vspace{-0.2in}
\end{figure*}

\noindent We now describe TripNet, an end to end reinforcement learning method for localizing and retrieving temporal activities in videos given a natural language query as shown in Figure ~\ref{fig:tripnet-model}.

{\bf Problem Formulation:}
The localization problem that we solve is defined as follows: Given an untrimmed video $V$ and a language query $L$, our goal is to temporally localize the specific clip $W$ in $V$ which is described by $L$. In other words, let us denote the untrimmed video as $V = \{f_n\}^N_{n=1}$ where $N$ is the number of frames in the video, we want to find the clip $W = \{f_{n}, \ldots, f_{n+k}\}$ that corresponds best to $L$. It is possible to solve this problem efficiently because videos have an inherent temporal structure, such that an observation made at frame $n$ conveys information about frames both in the past and in the future. Some challenges of the problem are, how to encode the uncertainty in the location of the target event in a video, and how to update the uncertainty from successive observations. While a Bayesian formulation could be employed, the measurement and update model would need to be learned and supervision for this is not available. 

Since it is computationally feasible to simulate the search process (in fact it is only a one-dimensional space, in contrast to standard navigation tasks) we adopt an reinforcement learning (RL) approach. We are motivated by human annotators who observe a short clip and make a decision to skip forward or backward in the video by some number of frames, until they can narrow down to the target clip. We emulate this sequential decision process using RL. Using RL we train an agent that can steer a fixed sized window around the video to find $W$ without looking at all frames of $V$. We employ the actor-critic method A3C~\cite{mnih2016asynchronous} to learn the policy $\pi$ that maps $(V, L) \to W$. The intuition is that the agent will take large jumps around the video until it finds visual features that identify proximity to $L$, and then it will start to take smaller steps as it narrows in on the target clip. 

\subsection{State and Action Space}
At each time step, the agent observes the current state, which consists of the sentence $L$ and a candidate clip of the video. The clip is defined by a bounding window [$W_{start}$, $W_{end}$] where start and end are frame numbers. At time step $t=0$, the bounding window is set to [$0$, $X$], where X is the average length of annotated clips within the dataset.\footnote{Note that this means that the search process always starts at the beginning of the video. We also explored starting at the middle and end of the clip and found empirically that starting at the beginning performed the best.} This window size is fixed and does not change. At each time step the state processing module creates a state representation vector which is fed into the policy learning module, on which it generates an action policy. This policy is a distribution over all the possible actions. An action is then sampled according to the policy. 

Our action space consists of 7 predefined actions: move the entire bounding window $W_{start}$, $W_{end}$ forward or backward by $h$ frames, $j$ frames, or 1 second of frames or TERMINATE. Where $h = N/10$ and $j = N/5$. These navigation steps makes aligning the visual and text features easier. Making the RL agent explicitly learn the window size significantly increases the state space and leads to a drop in accuracy and efficiency with our current framework. However, an additional function to adjust the box width over the fixed window size could be learned, analogous to the refinement of the anchor boxes used in object detectors.  These actions were chosen so that amount of movement is proportional to the length of the video. If the bounding window is at the start or end of the full video and an action is chosen that would push the bounding window outside the video's length, the bounding window remains the same as the previous time step. The action TERMINATE ends the search and returns the clip of the current state as the video clip predicted to be best matched to $L$. 

\subsection{TripNet architecture}
We now describe the architecture of TripNet, which is illustrated in Figure \ref{fig:tripnet-model}. 
\paragraph{State Processing Module}
At each time step, the state-processing module takes the current state as an input and outputs a joint-representation of the input video clip and the sentence query $L$. The joint representation is used by the policy learner to create an action policy from which the optimal action to take is sampled. The clip is fed into C3D \cite{tran2015learning} to extract the spatio-temporal features from the fifth convolutional layer. We mean-pool the C3D features across frames and denote the result as $x_M$.

To encode the sentence query, we first pass $L$ through a Gated Recurrent Unit (GRU)~\cite{chung2014empirical} which outputs a vector $x_L$. We then transform the query embedding into an attention vector that we can apply to the video embedding. To do so, the sentence query embedding $x_L$ is passed through a fully connected linear layer with a sigmoid activation function. The output of this layer is expanded to be the same dimension of $x_M$. We call the output of the linear layer the attention vector, $att_L$. We perform a Hadamard multiplication between $att_L$ and $x_M$ and the result is output as the state representation $s_t$. This attention unit is our gated attention architecture for activity localization. Hence, the gated attention unit is designed to gate specific filters based on the attention vector from the language query~\cite{dhingra2016gated}. The gating mechanism allows the model to focus on specific filters that can attend to specific objects, and their attributes, from the query description. 

In our experiments, we implement an additional baseline using a simple concatenation operation between the video and text representations to demonstrate the effectiveness of our gated-attention architecture. Here, we only perform self-attention over the mean pooled C3D features. Then, we take a Skip-Thought~\cite{kiros2015skip} encoding of the sentence query and concatenate it with the features of the video frames to produce the state representation. In the experiments, we denote this method as TripNet-Concat.

\paragraph{Policy Learning Module}
We use an actor-critic method to model the sequential decision process of grounding the language query to a temporal video location. The module employs a deep neural network to learn the policy and value functions. The network consists of a fully connected linear layer followed by an LSTM, which is followed by a fully connected layer to output the value function $v(s_t|\theta_v)$ and fully connected layer to output the policy $\pi(a_t | s_t,\theta_\pi)$, for state, $s_t$ and action, $a_t$ at time $t$, $\theta_v$ is the critic branch parameters and $\theta_\pi$ is actor branch parameters. The policy, $\pi$, is a probabilistic distribution over all possible actions given the current state. Since we are trying to model a sequential problem we use an LSTM so that the system can have memory of the previous states which will inevitably positively impact the future actions. Specifically, we use the asynchronous actor-critic method known as A3C \cite{mnih2016asynchronous} with Generalized Advantage Estimation~\cite{schulman2015high} that reduces policy gradient variance. The method runs multiple parallel threads that each run their own episodes and update global network parameters at the end of the episode.

Since the goal is to learn a policy that returns the best matching clip, we want to reward actions that bring the bounding windows [$W_{start}$, $W_{end}$] closer to the bounds of the ground truth clip. Hence, the action to take, should return a state that has a clip with more overlap with the ground-truth than the previous state. Therefore, we use reward shaping by having our reward be the difference of potentials between the previous state and current state. However, we want to ensure the agent is taking an efficient number of jumps and not excessively sampling the clip. In order to encourage this behavior, we give a small negative reward in proportion with the total number of steps thus far. As a result, the agent is encouraged to find the clip window as quickly as possible. We experiment to find the optimal negative reward factor $\beta$. We found using a negative reward factor results in the agent taking more actions with larger frame jump. Hence, our reward at any time step $t$ is calculated as follows:
\small 
\begin{equation}
reward_t = (IOU_t(s_t) - IOU_t(s_{t-1})) - \beta*t
\end{equation}
\normalsize
where we set $\beta$ to .01. We calculate the IOU between the clip of the state at time $t$, [$W_{start}^t$, $W_{end}^t$], and the ground truth clip for sentence $L$, [$G_{start}$, $G_{end}$] as follows:
\small
\begin{equation}
\label{iou}
IOU_t = \frac{min(W_{end}^t,G_{end})-max(W_{start}^t,G_{start})}{max(W_{end}^t,G_{end})-min(W_{start}^t,G_{start})}
\end{equation}
\normalsize

We use the common loss functions for A3C for the value and policy loss. For training the value function, we set the value loss to the mean squared loss between the discounted reward sum and the estimated value:
\small
\begin{equation}
Loss_{value} = \sum_t (R_t- v(s_t|\theta_v))^2 * \gamma_1,
\end{equation}
\normalsize
where we set $\gamma_1$ to .5 and $R_t$ is the accumulated reward. For training the policy function, we use the policy gradient loss:

\small
\begin{equation}
  \begin{aligned}
    Loss_{policy} =-\sum_t log(\pi(a_t|s_t,\theta_\pi)) *GAE(s_t) \\- \gamma_0*H(\pi(a_t|s_t,\theta_\pi)),
  \end{aligned}
\end{equation}
\normalsize 
where GAE is the generalized advantage estimation function, H is the calculation of entropy and $\gamma_0$ is set to 0.5.
Therefore, the total loss for our policy learning module is:
\small 
\begin{equation}
Loss = Loss_\pi + \gamma_1 + Loss_v.
\end{equation}
\normalsize
\section{Evaluation}
\label{sec:eval}
We evaluate the TripNet architecture on three video datasets, Charades-STA~\cite{gao2017tall}, ActivityNet Captions~\cite{krishna2017dense} and TACoS~\cite{regneri2013grounding}. Charades-STA was created for the moment retrieval task and the other datasets were created for video captioning, but are often used to evaluate the moment retrieval task. Note that we chose not to include the DiDeMo~\cite{hendricks2017localizing} dataset because the evaluation is based on splitting the video into 21 pre-defined segments, instead of utilizing continuously-variable start and end times. This would require a change in the set of actions for our agent. We do, however, compare our approach against the method from ~\cite{hendricks2017localizing} on other datasets. Please refer to supplementary materials for the details on the datasets as well as the implementation details of our system. 

\subsection{Experiments}
\label{sec:exp}

\setlength{\tabcolsep}{5pt}
\begin{table*}[t]
\begin{center}
\resizebox{\textwidth}{!}{
\begin{tabular}{l c ccc c ccc c ccc}
\toprule
                & & \multicolumn{3}{c}{ActivityNet}    & & \multicolumn{3}{c}{TACoS}   & & \multicolumn{3}{c}{Charades}                                                                                                                                         \\\cmidrule(l{2pt}r{2pt}){3-5} \cmidrule(l{2pt}r{2pt}){7-9} \cmidrule(l{2pt}r{2pt}){10-12} 
                
Method         && 
\multicolumn{1}{c}{\scriptsize $\alpha@.3$}& \multicolumn{1}{c}{\scriptsize $\alpha@.5$}& \multicolumn{1}{c}{\scriptsize $\alpha@.7$}&& \multicolumn{1}{c}{\scriptsize $\alpha@.3$}& \multicolumn{1}{c}{\scriptsize $\alpha@.5$}& \multicolumn{1}{c}{\scriptsize $\alpha@.7$}&& \multicolumn{1}{c}{\scriptsize $\alpha@.3$}& \multicolumn{1}{c}{\scriptsize $\alpha@.5$}& \multicolumn{1}{c}{\scriptsize $\alpha@.7$}   \\ 
\toprule
MCN~\cite{hendricks2017localizing}      &&  21.37 & 9.58 & - && 1.64 & 1.25 & - && - & - & -\\
CTRL~\cite{gao2017tall}      &&  28.70 & 14.00 & - && 18.32 & 13.3 & - && - & 23.63 & 8.89\\
TGN~\cite{chen2018temporally}       &&  45.51 & 28.47 & - && 21.77 & 18.9 & - && - & - & -\\
ABLR~\cite{yuan2018find}       && \textbf{55.67} & \textbf{36.79} & -&& - & - & - && - & - & -\\
ACRN~\cite{liu2018attentive}       &&  31.29 & 16.17 & - &&19.52 & 14.62 & - && - & - & -\\
MLVI~\cite{xu2018text}       &&  45.30 & 27.70 & 13.60 && - & - & - && \textbf{54.70} & 35.60 & 15.80\\
VAL \cite{song2018val} && - & - & - && 19.76 & 14.74 & - && - & 23.12 & 9.16\\
SM-RL\cite{wang2019language} && - & - & - && 20.25 & 15.95 & - && 24.36 & 11.17 & -\\
\midrule
TripNet-Concat   &&  36.75 & 25.64 & 10.25 && 18.24 & 14.16 & 6.47 && 41.84 & 27.23 & 12.62\\ 
TripNet-GA   &&  48.42 & 32.19 & \textbf{13.93} && \textbf{23.95} & \textbf{19.17} & \textbf{9.52} && 54.64 & \textbf{38.29} & \textbf{16.07}\\
\bottomrule
\end{tabular}
}
\caption{The accuracy of each method on ActivityNet, TACoS and Charades measured by IoU at multiple $\alpha$ values.}
\label{tab:acc}
\end{center}
\end{table*}

\paragraph{Evaluation Metric.} We use Intersection over Union (IoU) at different alpha thresholds to measure the difference between the ground truth clip and the clip that TripNet temporally aligns to the query. If a predicted window and the ground truth window have an IoU that is above the set alpha threshold, we classify the prediction as correct, otherwise we classify it as incorrect. See Equation \ref{iou} for how the IoU is calculated. The results over the different datasets and methods are shown in Table \ref{tab:acc}. Most previous work uses a R@k-IoU, which is a IoU score for the top k returned clips. These works used a sliding window approach allowing their alignment system to return k-top candidate windows based on confidence scores. In contrast, TripNet searches the video until it finds the best aligned clip and returns only that clip. As a result, all reported IoU scores are measured at R@1.

\subsubsection{Comparison and Baseline Methods}
We compare against other methods both from prior work and a baseline version of the TripNet architecture. The prior work we compare against is as follows: MCN~\cite{hendricks2017localizing}, CTRL~\cite{gao2017tall}, TGN~\cite{chen2018temporally}, ABLR~\cite{yuan2018find}, ACRN~\cite{liu2018attentive}, MLVI~\cite{xu2018text}, VAL~\cite{song2018val} and SM-RL~\cite{wang2019language}. Besides SM-RL, all prior works tackle the task by learning to jointly represent the ground truth moment clip and the moment description query. To generate candidate windows during testing, these methods go over the whole video using a sliding window and then, choose the candidate window that best corresponds to the query encoding. This methodology relies on seeing all frames of the video at least once, if not more, during test time.

We provide accuracy results in Table \ref{tab:acc} for 3 datasets, compiling the performance numbers reported in the prior works for comparison to the performance of TripNet. We can see that, in terms of accuracy, TripNet outperforms all other methods on the Charades-STA and TACoS datasets, and that it performs comparably to the state of the art on ActivityNet Captions. Using a state processing module for TripNet that does not use attention (TripNet-Concat) performs consistently worse than the state processing module for TripNet that uses the gated attention architecture (TripNet-GA), since multi-modal fusion between vision and language shows improvement with different attention mechanisms.

\small
\begin{table}[t!]
  \begin{center}
  \footnotesize
    \resizebox{\textwidth}{!}{
    \begin{tabular}{l c c c}
      \toprule
      \textbf{Dataset} & \% of Frames Used  & Avg. \# Actions & Bounding Window Size\\
      \toprule
        ActivityNet           & 41.65 & 5.56 & 35.7s\\
        Charades                   & 33.11 & 4.16 & 8.3s\\
        TACoS                          & 32.7 & 9.84 & 8.0s\\
      \toprule
    \end{tabular}
    }
    \vspace{0.1in}
    \caption{The efficiency of the TripNet architecture on different datasets. Number of actions includes the TERMINATE action. Size of the bounding window is in seconds. The methods were tested on the same Titan Xp GPU.}
    \label{tab:tripnet-eff}
    \vspace{-0.2in}
  \end{center}
\end{table}
\normalsize

\small  
\begin{table}
  \begin{center}
  \footnotesize
  \resizebox{.6\textwidth}{!}{
    \begin{tabular}{l c c c}
      \toprule
      Method & Charades & ActivityNet & TACoS\\
      \toprule
      CTRL \cite{gao2017tall}           & 44.19ms & 218.95ms & 342.12ms\\
      MCN \cite{hendricks2017localizing} & 78.16ms & 499.32ms & 674.01ms\\
      TGN \cite{chen2018temporally} & 18.2ms & 90.2ms & 144.7ms\\
      TripNet-GA                        & 5.13ms & 6.23ms & 11.27ms\\ 
      \toprule
    \end{tabular}
    }
    \vspace{0.1in}
    \caption{The average time in milliseconds that it takes to localize a moment on different datasets.}
    \label{tab:time}
    \vspace{-0.2in}
  \end{center}
\end{table}
\normalsize 
 
More specifically, in Table \ref{tab:acc}, ABLR processes all frames and encodes the intermediate representations of each frame and word using Bi-LSTM, followed by two levels of attention over it. It then regresses the coordinates over this comprehensive intermediate representation. Hence, this regression-based approach over all frames is beneficial for the TALL task. However, it is computationally inefficient to perform this processing over all frames. On the other hand, in our method feature extraction is only performed if requested by the RL step and is suitable for long surveillance videos where ABLR may not be practical. We also see an overall drop in performance of TripNet on TACoS, despite outperforming other recent work. We don't believe this is due to the length of the videos but instead due to the nature of the content in the video. TACoS contains long cooking videos all taken in the same kitchen making the dataset more difficult. While ActivityNet videos are up to four times longer than Charades-STA videos, TripNet is nonetheless able to maintain high accuracy on ActivityNet, illustrating TripNet's ability to scale to videos of different lengths.

In the analysis of our results, we found that one source of IoU inaccuracy was the size of the bounding window. TripNet currently uses a fixed size bounding window that the agent moves around the video until it returns a predicted clip. The size of the fixed window is equal to the mean length of the ground truth annotated clips. A possible direction for future work would be to add actions that expand and contract the size of the bounding window. 

\subsubsection{Efficiency} 
We now describe how the different competing methods construct their candidate windows, in order to provide an understanding of the computational cost of each approach. 
\noindent MCN \cite{hendricks2017localizing}: creates 21 candidate windows across the video. 
CTRL \cite{gao2017tall}, VAL \cite{song2018val}, ACRN \cite{liu2018attentive}: use a sliding windows of multiple sizes to generate candidate windows.
TGN \cite{chen2018temporally}: does not use a sliding window but generates candidate windows across the video at different scales.
ABLR\cite{yuan2018find}: encodes the entire video and then uses attention via the language query to localize the clip. The method still requires two passes over the entire video.
MLVI \cite{xu2018text}: trains a separate network to generate candidate window proposals.
\textbf{TripNet}: TripNet is the only method that does not need to watch the entire video to temporally localize a described moment. Instead, our trained agent efficiently moves a candidate window around the video until it localizes the described clip. Measurements of efficiency are described in Table \ref{tab:tripnet-eff}. In order to get a better understanding of efficiency, we run the CTRL ~\cite{gao2017tall}, TGN \cite{chen2018temporally}, MCN \cite{hendricks2017localizing}, and TripNet-GA over the dataset test splits and compute the average time it takes to localize a moment in the video. This is shown in Table~\ref{tab:time}.

\subsubsection{Qualitative Results}
\begin{figure*}[h!]
    \centering
    \includegraphics[width=\textwidth]{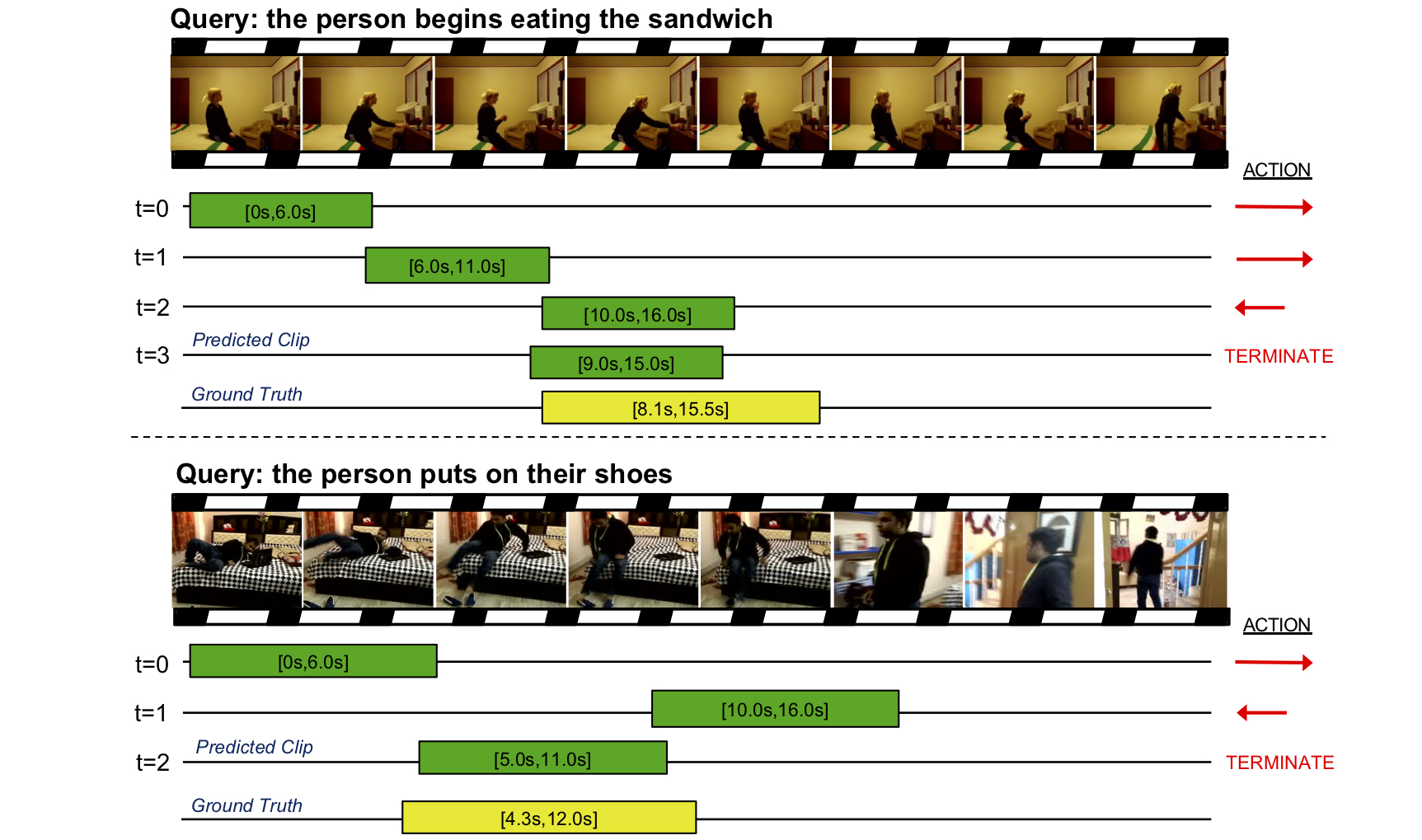}
    \caption{Qualitative performance of TripNet-GA: We show two examples where the TripNet agent skips through the video looking at different candidate windows before terminating the search. Both these videos are from the Charades-STA dataset.}
    \label{fig:qualresult}
    \vspace{-0.1in}
\end{figure*}

In Figure \ref{fig:qualresult}, we show the qualitative results of TripNet-GA on the Charades-STA dataset. In the figure, we show the sequential list of actions the agent takes in order to temporally localize a moment in the video. The green boxes represent the bounding window of the state at time t and the yellow box represents the ground truth bounding window. The first video is 33 seconds long (792 frames) and the second video is 20 seconds long (480 frames). In both examples the agent skips both backwards and forwards. In the first example, TripNet sees 408 of the frames of the video, which is 51\% of the video. In the second example TripNet sees 384 frames of the video, which is 80\% of the frames. 
\section{Conclusion}
\label{sec:conclusion}
Localizing moments in long, untrimmed videos using natural language queries is a useful and challenging task for fine-grained video retrieval. Most previous works are designed to analyze 100\% of the video frames. In this paper, we have introduced a system that uses a gated-attention mechanism over cross-modal features to automatically localize a moment in time given a natural language text query with high accuracy. Our model incorporates a policy network which is trained for efficient search performance, resulting in an  system that on average examines less then 50\% of the video frames in order to localize a clip of interest, while achieving a high level of accuracy.
\bibliography{egbib}
\begin{appendices}
\section{State Processing Module - Gated Attention}
At each time step, the state-processing module takes the current state as an input and outputs a joint-representation of the input video clip and the sentence query $L$. The joint representation is used by the policy learner to create an action policy over which the optimal action to take, is sampled. The clip is fed into C3D \cite{tran2015learning} to extract the spatio-temporal features from the fifth convolutional layer. We mean-pool the C3D features across frames and denote the result as $x_M$. To encode the sentence query, we first pass $L$ through a Gated Recurrent Unit (GRU)~\cite{chung2014empirical} which outputs a vector $x_L$. 
TripNet then uses a gated attention architecture to combine the language and video modalities. 

\begin{figure}[t]
    \centering
    \includegraphics[width=3.2in]{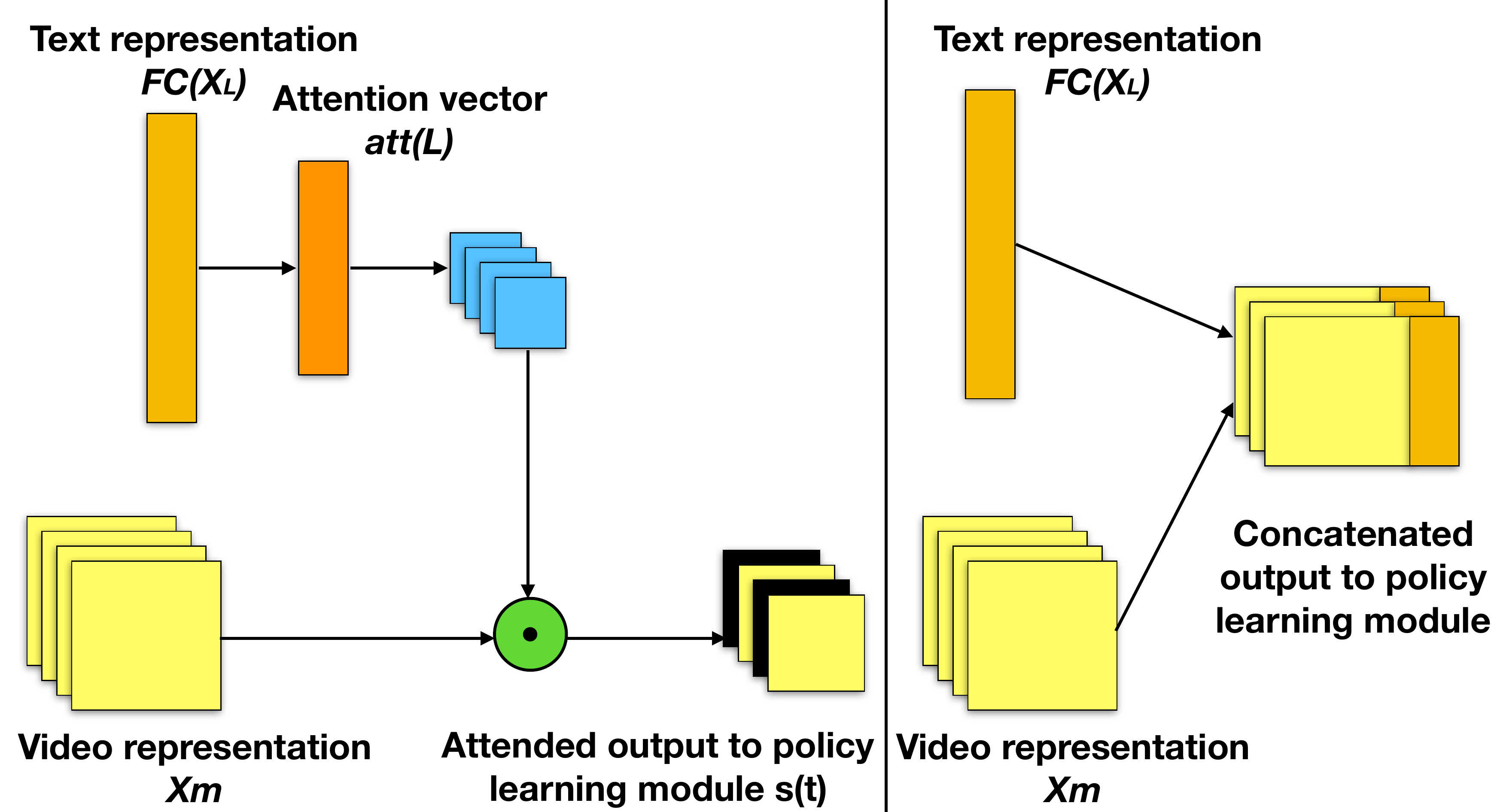}
    \caption{\small{\bf This figure shows on the left TripNet-GA and on the right TripNet-Concat, where gated-attention over text features and simple concatenation are explored, respectively.}}
    \label{fig:attvsconcat}
    \vspace{-0.1in}
\end{figure}

In addition to comparing against prior works, we run TripNet without the gated attention mechanism, shown as TripNet-Concat in Figure \ref{fig:attvsconcat}. TripNet-Concat does self attention over the mean pooled C3D features of the video frames and concatenates the output with the Skip-Thought~\cite{kiros2015skip} encoding of the sentence query to produce the state representation. Testing this method allows us to explore the performance of the state processing module separately from the policy learning module. The difference between the gated-attention and concatenation methods are illustrated in Figure \ref{fig:attvsconcat}.

\section{Datasets}
We evaluate the TripNet architecture over three video datasets, Charades-STA~\cite{gao2017tall}, ActivityNet Captions~\cite{krishna2017dense} and TACoS~\cite{regneri2013grounding}. Charades-STA was created specifically for the moment retrieval task and the other datasets were created for the video captioning task but are commonly used to evaluate the moment retrieval task. Note that we chose not to include the DiDeMo~\cite{hendricks2017localizing} dataset because in previous work, the evaluation is based off splitting the video into 21 pre-defined segments, instead of specific start and end times. This would mean changing the set of actions for our agent and we wanted the set of actions to be consistent across datasets. We do, however, compare against the method from ~\cite{hendricks2017localizing} on other datasets. 

All the datasets that we use contain untrimmed videos and natural language descriptions of specific moments in the videos. These language descriptions are annotated with the corresponding start and end time of the corresponding clip. \\

\noindent\textbf{Charades-STA~\cite{gao2017tall}}. This dataset takes the original Charades dataset, which contains video annotations of activities and video descriptions, and transforms these annotations to temporal sentence annotations which have a start and end time. This dataset was made for the task of temporal activity localization based on sentence descriptions. There are 13898 video to sentence pairs in the dataset. For evaluation, we use the dataset's predefined test and train splits. On average, the videos are 31 seconds long and the described temporally annotated clips are 8 seconds long. 

\textbf{ActivityNet Captions~\cite{krishna2017dense}}. In order to test the robustness of our system with longer video lengths, we use ActivityNet Captions that contains 100K temporal description annotations over 20k videos. This dataset was originally created for video captioning but is easily adaptable to our task and showcases the efficient performance of our architecture on longer videos. On average, the videos are 2.5 minutes long and the described temporally annotated clips are 36 seconds long. 

\textbf{TACoS~\cite{regneri2013grounding}}. This dataset contains both activity labels and natural language descriptions, both with temporal annotations for 127 videos. Following previous work, for evaluation we randomly split the dataset into 50\% for training, 25\% for validation and 25\% for testing. We choose this dataset because of its long videos, which are 4.5 minutes long on average and the temporally annotated clips are 5 seconds long on average.\\

\section{Implementation details.}
During training, we take a video and a single query sentence that has a ground truth temporal alignment in the clip. At time $t=0$ we set the bounding window [$W_{start}$, $W_{end}$] to be [0, X] where X is the average length of ground truth clips in the dataset. This means that this is the initial clip in the sequential decision process. Furthermore, it also means that the first actions selected will most likely be skipping forward in the video. The input to the system is $X$ sequential video frames and a sentence query. The sentence is first encoded through a Gated Recurrent Unit of size 256 and then through a fully-connected linear layer of size 512 with sigmoid activation. We run the video frames within the bounding window through a 3D-CNN \cite{tran2014c3d} which is pre-trained on the Sports-1M dataset and extract the 5th convolution layer. The A3C reinforcement learning method is then used for the policy learning module and is trained with stochastic gradient descent (SGD) with a learning rate of $.0005$. The first fully-connected (FC) layer of the policy learning module is $256$ dimensions and is followed by an long short term memory (LSTM) layer of size $256$. During training, we set A3C to run 8 parallel threads.

\end{appendices}
\end{document}